%% file: main.tex
\documentclass{article}

\PassOptionsToPackage{round, sort&compress}{natbib}
\usepackage[final]{nips_2018}

\usepackage[utf8]{inputenc} % allow utf-8 input
\usepackage[T1]{fontenc}    % use 8-bit T1 fonts
\usepackage{hyperref}       % hyperlinks
\usepackage{url}            % simple URL typesetting
\usepackage{booktabs}       % professional-quality tables
\usepackage{amsfonts}       % blackboard math symbols
\usepackage{nicefrac}       % compact symbols for 1/2, etc.
\usepackage{microtype}      % microtypography

\usepackage{mathtools}
\usepackage{amssymb}
\usepackage{amsthm}
\usepackage{paralist}
\usepackage{subcaption}
\usepackage{multirow}
\usepackage{wrapfig}
\usepackage[dvipsnames]{xcolor}
\usepackage[ruled,vlined]{algorithm2e}
\usepackage[nameinlink,capitalise]{cleveref}
\usepackage[acronym,smallcaps,nowarn,section,nogroupskip,nonumberlist]{glossaries}
\usepackage{todonotes}
\usepackage{titlesec}
\usepackage{siunitx}

\definecolor{red}{HTML}{E41A1C}
\definecolor{orange}{HTML}{FF7F00}
\definecolor{yellow}{HTML}{FFC020}
\definecolor{green}{HTML}{4DAF4A}
\definecolor{blue}{HTML}{377EB8}
\definecolor{purple}{HTML}{984EA3}
\graphicspath{{images/}}
\hypersetup{%
  colorlinks,
  linkcolor={violet!70!black},
  citecolor={YellowOrange!70!black},
  urlcolor={Aquamarine!70!black}
}
\Crefname{algocf}{Algorithm}{Algorithms}
\crefname{algorithm}{Algorithm}{Algorithms}
\crefname{figure}{Figure}{Figure}
\crefname{table}{Table}{Table}
\crefname{section}{§}{§§}
\Crefname{section}{§}{§§}
\crefformat{equation}{(#2#1#3)}
\glsdisablehyper{}
\newacronym{ML}{ml}{machine learning}
\newacronym{CCA}{cca}{canonical correlation analysis}
\newacronym{VQA}{vqa}{visual question answering}
\newacronym{VD}{vd}{visual dialogue}
\newacronym{SGD}{sgd}{stochastic gradient descent}
\newacronym{SOTA}{sota}{state-of-the-art}
\presetkeys{todonotes}{%
  backgroundcolor=blue!10!white,
  linecolor=blue!10!white,
  bordercolor=blue!10!white
}{}

\newcommand{\bfx}{\mathbf{x}}

\makeatletter
\renewcommand{\@noticestring}{}
\makeatother

\title{Visual Dialogue without Vision or Dialogue}

\author{
  Daniela Massiceti\thanks{Equal Contribution} ~~~~ Puneet K. Dokania$^*$ ~~~ N. Siddharth$^*$ ~~~ Philip H.S. Torr \\
  University of Oxford \\
  \texttt{\{daniela, puneet, nsid, phst\}@robots.ox.ac.uk}
}

\newcommand{\norm}[1]{\left\lVert#1\right\rVert}

\setdefaultleftmargin{4ex}{2ex}{}{}{}{}

\input{maths-basic}             % import extra maths
\newcommand\numberthis{\addtocounter{equation}{1}\tag{\theequation}}
\newcommand{\rank}[2][black]{\textcolor{#1}{\textcircled{\scalebox{0.6}{#2}}}}

\titlespacing\section{0pt}{3pt plus 4pt minus 2pt}{0pt plus 2pt minus 2pt}
\titlespacing\subsection{0pt}{3pt plus 4pt minus 2pt}{0pt plus 2pt minus 2pt}

\begin{document}
\maketitle

\vspace*{-1.6\baselineskip}
\begin{abstract}
\vspace*{-0.8\baselineskip}
  We characterise some of the quirks and shortcomings in the exploration of \gls{VD}---a sequential question-answering task where the questions and corresponding answers are related through given visual stimuli.
  To do so, we develop an embarrassingly simple method based on \gls{CCA} that, on the standard dataset, achieves near state-of-the-art performance on mean rank (MR).
  In direct contrast to current complex and over-parametrised architectures that are both compute and time intensive, our method \emph{ignores the visual stimuli}, \emph{ignores the sequencing of dialogue}, \emph{does not need gradients}, uses \emph{off-the-shelf} feature extractors, has at least an \emph{order of magnitude fewer parameters}, and learns in \emph{practically no time}.
  We argue that these results are indicative of issues in current approaches to \acrlong{VD} and conduct analyses to highlight implicit dataset biases and effects of over-constrained evaluation metrics.
  Our code is publicly available\footnote{\url{https://github.com/danielamassiceti/CCA-visualdialogue}}.
\end{abstract}

\glsresetall
\vspace*{-1.2\baselineskip}
\section{Introduction}
\vspace*{-0.5ex}
\label{sec:intro}

\begin{wrapfigure}[19]{r}{0.34\textwidth}
  \vspace*{-2.5\baselineskip}
  \scriptsize
  \textbf{Caption}: A man and a woman sit on the street in front of a large mural painting.\\[1pt]
  \begin{tabular}{@{}c@{}}
    \includegraphics[width=\linewidth]{hook.jpg} \\
    \begin{tabular}{@{}l@{\hspace*{8pt}}r@{}}
      \textbf{Question}               & \textbf{Answer} \\
      \cmidrule(r){1-1}\cmidrule{2-2}
      How old is the baby?            & About 2 years old \\
      What color is the remote?       & White \\
      Where is the train?             & On the road \\
      How many cows are there?        & Three \\
    \end{tabular}
  \end{tabular}
  \vspace*{-2ex}
  \caption[VD]{%
    \small%
    Failures in \emph{visual} dialogue.
    Visually-unrelated questions, and their visually-unrelated plausible answers\protect \footnotemark.
  }
  \label{fig:vd-fail}
\end{wrapfigure}
\footnotetext{From online demos of SOTA models--\href{http://demo.visualdialog.org/}{VisDial}~\citep{Das_VisualDialog} and \href{http://www.robots.ox.ac.uk/~daniela/research/flipdial/1vd_demo}{FlipDial}~\citep{massiceti2018}}

Recent years have seen a great deal of interest in \emph{conversational AI}, enabling natural language interaction between humans and machines, early pioneering efforts for which include ELIZA \citep{weizenbaum1966eliza} and SHRDLU \citep{winograd1971procedures}.
%
% SHRDLU even involved virtual interaction in an environment through (comparatively) rudimentary visual processing.
%
This resurgence of interest builds on the ubiquitous successes of neural-network-based approaches in the last decade, particularly in the perceptual domains of vision and language.

A particularly thriving sub-area of interest in conversational AI is that of \emph{visually grounded} dialogue, termed \gls{VD}, involving an AI agent conversing with a human about visual content \citep{Das_VisualDialog,visdial_rl,massiceti2018}.
Specifically, it involves answering questions about an image, given some dialogue history---a fragment of previous questions and answers.
Typical approaches for learning to do \gls{VD}, as is standard practice in \gls{ML}, involves defining an objective to achieve, procuring data with which to learn, and establishing a measure of success at the stated objective.

The objective for \gls{VD} is reasonably clear at first glance---answer in sequence, a set of questions about an image.
The primary choice of dataset, \textit{VisDial} \citep{Das_VisualDialog}, addresses precisely this criterion, involving a large set of images, each paired with a dialogue---a set of question-answer pairs---collected by pairs of human annotators playing a game to understand an image through dialogue.
And finally, evaluation measures on the objective are typically defined through some perceived value of a human-derived ``ground-truth'' answer in the system.

However, as we will demonstrate, certain quirks in the choices of the above factors, can lead to unintentional behaviour (c.f.\ \cref{fig:vd-fail}), which leverages implicit biases in data and methods, to potentially misdirect progress from the desired objectives.
Intriguingly, we find that in contrast to \gls{SOTA} approaches that employ complex neural-network architectures using complicated training schemes over millions of parameters and taking many hours of time and expensive GPU compute resources, the simple \gls{CCA}-based method only uses standard off-the-shelf feature extractors, avoids computing gradients, involves a few hundred thousand parameters and requires just a few seconds on a CPU to achieve comparable performance on the mean rank (MR) metric---all \emph{without requiring the image or prior dialogue!}

\section{(Multi-View) \acrshort{CCA} for \acrshort{VD}}
\label{sec:cca}

We begin with a brief preliminary for \gls{CCA}~\citep{hotelling1936} and its multi-view extension~\citep{kettenringMultiCCA1971}.
In (standard 2-view) \gls{CCA}, given access to paired observations $\{\bfx_1 \in \mathbb{R}^{n_1 \times 1}, \bfx_2 \in \mathbb{R}^{n_2\times 1}\}$, the objective is to jointly learn projection matrices \( W_1 \in \mathbb{R}^{n_1 \times p} \) and \( W_2 \in \mathbb{R}^{n_2 \times p} \) where \( p \leq \min (n_1,n_2) \), that maximise the correlation between the projections, formally \(\mathrm{corr}(W_1^{\top} \bfx_1, W_2^{\top} \bfx_2)\).

Multi-view \gls{CCA}, a generalisation of \gls{CCA}, extends this to associated data across~\(m\) domains, learning projections \( W_i \in \mathbb{R}^{n_i \times p}, \: i \in \{1, \dots, m\}\).
\citet{kettenringMultiCCA1971} shows that \(W_i\) can be learnt by minimising the Forbenius norm between each pair of views, with additional constraints over the projection matrices~\citep{hardoonMultiCCA2004}.
Optimising the multi-view \gls{CCA} objective then reduces to solving a generalized eigenvalue decomposition problem, \( Av = \lambda Bv \), where $A$ and $B$ are derived from the inter- and intra-view correlation matrices (c.f.\ \cref{sec:app-cca})~\citep{bachkernelICA2002}.

Projection matrices~\(W_i\) are extracted from corresponding rows (for view~\(i\)) and the top~\(p\) columns of the (eigenvalue sorted) eigenvector matrix corresponding to this eigen-decomposition.
A sample~\(\bfx_i\) from view~\(i\) is then embedded as \(\phi_{q}(\bfx_i, W_i) = (W_i\, D^q_p)^{\!\top} \bfx_i\), where \(D^q_p = \mathrm{diag}(\lambda_1^q, \cdots, \lambda_p^q)\) and $\lambda_1 \geq \cdots \geq \lambda_p$ are the eigenvalues.
A scaling, $q \in \mathbb{R}$, controls the extent of eigenvalue weighting, reducing to the standard objective at \(q = 0\)\footnote{There are cases where values of $q > 0$ have been shown to give better performance~\citep{gongmultiCCA14}.}.
With this simple objective, one can tackle a variety of tasks at test time---ranking and retrieval across all possible combinations of multiple views---where the cosine similarity between (centred) embedding vectors captures correlation.

For \gls{VD}, given a dataset of images~\(I\) and associated question-answer (\(Q\text{-}A\)) pairs, joint embeddings between question and answer (and optionally, the image) are learnt, with projection matrices~\(W_Q, W_A, (\text{and\,} W_I)\), as appropriate.
At test time, correlations can be computed between any, and all, combinations of inputs, helping measure suitability against the desired response.

\section{Experimental Analyses}
\label{sec:experiments}

In order to employ \gls{CCA} for \gls{VD}, we begin by transforming the input images~\(I\), questions~\(Q\), and answers~\(A\), into lower-dimensional feature spaces.
For the images, we employ the standard pre-trained ResNet34~\citep{He_2016deep} architecture, extracting a 512-dimensional feature---the output of the \emph{avg pool} layer after \emph{conv5}.
For the questions and answers, we employ the FastText~\citep{bojanowski2017enriching} network to extract 300-dimensional embeddings for each of the words.
We then simply average the embeddings~\citep{aurora2017simple} for the words, with suitable padding or truncation (up to a maximum of 16 words), to obtain a 300-dimensional embedding for the question or answer.

\setlength{\columnsep}{5pt}%
\begin{wraptable}[9]{r}{0.36\textwidth}
  \vspace*{-1.2\baselineskip}
  \caption{\small \gls{CCA} vs.\ \gls{SOTA}: number of learnable parameters and training time.}
  \vspace*{-1.5ex}
  \label{tab:compute-time}
  \defcitealias{lu2017best}{HCIAE-G-DIS}
  \defcitealias{Das_VisualDialog}{VisDial}
  \defcitealias{massiceti2018}{FlipDial}
  \centering
  \scalebox{0.88}{%
  \footnotesize
  \begin{tabular}{@{}lS[table-format=1.2e1]@{\,}S[table-format=1.1e1]@{}}
    \toprule
    {Model}                          & {\#Params}      & {Train time (s)} \\
    \midrule
    \citetalias{lu2017best}          & 2.12e7          & {--}   \\
    \citetalias{Das_VisualDialog}    & 2.42e7          & {--}   \\
    \citetalias{massiceti2018}       & 1.70e7          & 2.0e5  \\
    \cmidrule{2-3}
    \gls{CCA} (A-Q)                  & 1.80e5          & 2.0e0  \\
    \midrule
    \textbf{Factor (\(\approx\))}
    & \multicolumn{1}{c}{\textbf{90}}
    & \multicolumn{1}{c}{\textbf{10}\(^{\mathbf{5}}\)} \\
    \bottomrule
  \end{tabular}}
\end{wraptable}
We then set the hyper-parameters for the \gls{CCA} objective as \(p = 300, q = 1\), based off of a simple grid search over feasible values, such that we learn a 300-dimensional embedding space that captures the correlations between the relevant domains.
It is important to note that the \gls{SOTA} approaches~\citep{Das_VisualDialog,visdial_rl,massiceti2018} also employ pre-trained feature extractors---the crucial difference between approaches is the complexities in modelling and computation \emph{on top of} such feature extraction, as starkly indicated in \cref{tab:compute-time}.

We then learn two joint embeddings---between just the answers and questions, denoted A-Q, and between the answers, questions, and images, denoted A-QI.
Note that the answer is always present, since the stipulated task in \gls{VD} is to answer a given question.
The first allows us to explore the utility (or lack thereof) of the image in performing the \gls{VD} task.
The second serves as a useful indicator of how unique any question-image pairing is, in how it affects the ability to answer---performance closer to that of A-Q indicating fewer unique pairings.
Also, when embedding all three of A, Q, and I, at test time, we only employ Q to compute a match against a potential answer.

Having now learnt an embedding, we evaluate our performance using the standard ranking measure employed for the \textit{VisDial} dataset.
Here, for a given image and an associated question, the dataset provides a set of 100 candidate answers, which includes the human-derived ``ground-truth'' answer.
The task then, is to rank each of the 100 candidates, and observe the rank awarded to the ``ground-truth'' answer.
In our case, we rank on correlation, computed as the cosine distance between centered embeddings between the question and a candidate answer.
Then, for all the answers we compute the mean rank (MR), mean reciprocal rank (MRR) (inverse harmonic mean of rank), and recall at top 1, 5, and 10 candidates---measuring how often the ``ground-truth'' answer ranked within that range.

\setlength{\columnsep}{5pt}%
\begin{wraptable}[12]{r}{0.44\textwidth}
  \vspace*{-1\baselineskip}
  \caption{%
    \small%
    Results for \gls{SOTA} vs.\ \gls{CCA} on the \textit{VisDial} dataset.
    \gls{CCA} achieves comparable performance while ignoring both image and dialogue sequence.%
  }
  \label{tab:cca-analysis}
  \centering \footnotesize
  \newcommand{\rt}[2]{\parbox[t]{2mm}{\multirow{#1}{*}{\rotatebox[origin=c]{90}{#2}}}}
  \def\sota{\rt{3}{SotA}}
  \def\lcca{\rt{4}{\acrshort{CCA}}}
  \def\vold{\rt{2}{\emph{v0.9}}}
  \def\vnew{\rt{2}{\emph{v1.0}}}
  \defcitealias{lu2017best}{HCIAE-G-DIS}
  \defcitealias{wu2017you}{CoAtt-GAN}
  \defcitealias{Das_VisualDialog}{HREA-QIH-G}
  \def\hci{\citetalias{lu2017best}}
  \def\coa{\citetalias{wu2017you}}
  \def\hrea{\citetalias{Das_VisualDialog}}
  \def\s{\hspace*{5pt}}
  \scalebox{0.83}{%
  \begin{tabular}{@{}c@{\s}l@{\s}>{\footnotesize}l@{\s}l@{\s}l@{\s}l@{\s}l@{\s}l@{}}
    \toprule
          &       & Model     & MR     & R@1    & R@5    & R@10   & MRR     \\
    \midrule
    \sota & \vold & \hci      & {\bf 14.23} & 44.35  & 65.28  & 71.55  & 0.5467  \\
          &       & \coa      & {\bf 14.43} & 46.10  & 65.69  & 71.74  & 0.5578  \\
          &       & \hrea     & {\bf 16.79} & 42.28  & 62.33  & 68.17  & 0.5242  \\
    \midrule
    \lcca & \vold & A-Q       & {\bf 16.21} & 16.85 & 44.96 & 58.10 & 0.3041 \\
          &       & A-QI (Q)  & 18.27       & 12.24 & 35.55 & 50.88 & 0.2439 \\
    \cmidrule(l){2-8}
          & \vnew & A-Q       & 17.07 & 16.18 & 40.18 & 55.35 & 0.2845  \\
          &       & A-QI (Q)  & 19.25 & 12.63 & 32.88 & 48.68 & 0.2379  \\
    \bottomrule
  \end{tabular}}
\end{wraptable}

%
%% Analyse the quantitative results.
The results, in \cref{tab:cca-analysis}, show that the simple \gls{CCA} approach achieves comparable performance on the mean rank (MR) metric using the A-Q model that \emph{doesn't use the image or dialogue sequence!}
This solidifies the impression, from \cref{fig:vd-fail}, that there exist implicit correlations between just the questions and answers in the data, that can be leveraged to perform ``well'' on a task that simply requires matching ``ground-truth'' answers.
Our experiments indicate that for the given dataset and task, one need not employ anything more complicated than an exceedingly simple method such as \gls{CCA} on pre-trained feature extractors, to obtain plausible results.

\begin{wrapfigure}[15]{r}{0.44\linewidth}
  \vspace*{-1.7\baselineskip}
  \centering\tiny
  \begin{tabular}{@{}c@{\,\,}p{0.4\linewidth}@{\,}p{0.35\linewidth}@{}}
    \toprule
    Image
    & \shortstack{Question \\[-2pt] (Rank) GT Answer}
    & \shortstack{\acrshort{CCA} Top-3 \\[-2pt] (Rank) Answer} \\
    \midrule
    %% Example 1
    \multirow{6}{*}{\includegraphics[trim={0 2cm 0 3mm},clip,width=0.21\linewidth]{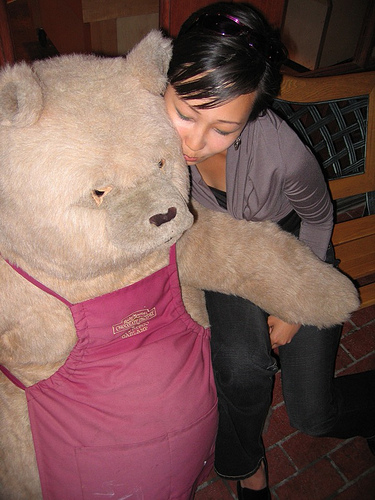}}
    %% Q1
    & What colour is the bear?
    & \rank{1} White and brown \\
    & \rank{51} Floral white & \rank{2} Brown and white \\
    && \rank{3} Brown, black \& white \\
    \cmidrule{2-3}
    %% Q2
    & Does she have long hair?
    & \rank{1} No, it is short hair \\
    & \rank{41} No & \rank{2} Short \\
    && \rank{3} No it's short \\[1ex]
    %% Example 2
    \multirow{6}{*}{\includegraphics[width=0.21\linewidth]{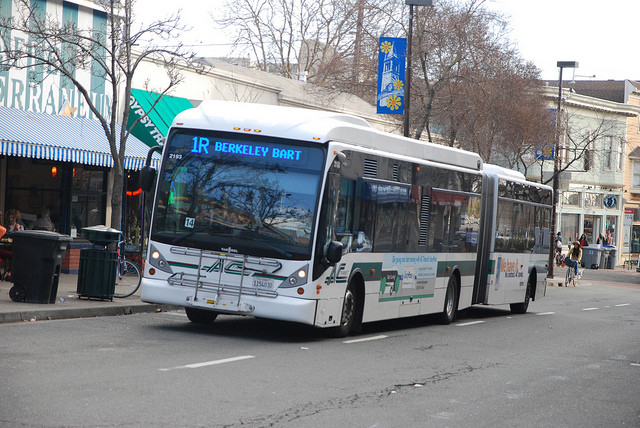}}
    %% Q1
    & Can you see any passengers?
    & \rank{1} No \\
    & \rank{48} Not really & \rank{2} Zero \\
    && \rank{3} No I can not \\
    \cmidrule{2-3}
    %% Q2
    & Are there people not on bus?
    & \rank{1} No people \\
    & \rank{22} Few & \rank{2} No, there are no people around \\
    && \rank{3} I don't see any people \\
    \bottomrule
  \end{tabular}
  \vspace*{-1ex}
  \caption{%
    \small%
    Qualitative results for the A-Q model showing the \emph{top-3} ranked answers for questions where the ground-truth answer is given a low rank---showing them to be perfectly feasible.
  }
  \label{fig:qualitative-examples}
\end{wrapfigure}
Moreover, another factor that needs to be considered, is that the evaluation metric itself, through the chosen task of candidate-answer ranking, can be insufficient to draw any \emph{actual} conclusions about how well questions were answered.
To see this, consider \cref{fig:qualitative-examples}, where we deliberately pick examples that rank the ``ground-truth'' answer poorly despite \gls{CCA}'s top-ranked answers all being plausible alternatives.
This clearly illustrates the limitations imposed by assuming a single ``ground-truth'' answer in capturing the breadth of correct answers.

To truly judge the validity of the top-ranked answers, regardless of ``ground-truth'' would require thorough human-subject evaluation.
However, as a cheaper, but heuristic alternative, we quantify the validity of the top answers, in relation to the ``ground truth'', using the correlations themselves.
For any given question and candidate set of answers, we cluster the answers based on an automatic binary thresholding (Otsu~\citep{otsu1979threshold}) of the correlation with the given question.
We then compute the following two statistics based on the threshold
\begin{inparaenum}[i)]
\item the average variance of the correlations in the lower-ranked split, and
\item the fraction of questions that have correlation with ``ground truth'' answer higher than the threshold.
\end{inparaenum}
The intention being that (i)~quantifies how closely clustered the top answers are, and (ii)~quantifies how often the ``ground-truth'' answer is in this cluster.
Low values for the former, and high values for the latter would indicate that there exists an equivalence class of answers, all relatively close to the ground-truth answer in terms of their ability to answer the question.
Our analysis for the \textit{VisDial v0.9} dataset reveals values of (i) 0.1180 and (ii) 86.95\%, supporting our claims that \gls{CCA} recovers plausible answers.

We note that the \textit{VisDial} dataset was recently updated to version 1.0, where the curators try to ameliorate some of the issues with the single-``ground-truth'' answer approach.
They incorporate a human-agreement scores for candidate answers, and introduce a modified evaluation which weighs the predicted rankings by these scores.
We include our performance on the (held-out) test set for \textit{VisDial v1.0} in the bottom row of \cref{tab:cca-analysis}.
However, in making this change, the primary evaluation for this data has now become an explicit classification task on the candidate answers\footnote{See \cref{sec:response} for an update on this characterisation.}---requiring access, at train time, to all (100) candidates for every question-image pair \citep[see Table 1, pg 8.][]{Das_VisualDialog} and the evaluation results of the \href{https://visualdialog.org/challenge/2018}{Visual Dialog Challenge 2018}.
For the stated goals of \gls{VD}, this change can be construed as unsuitable as it falls into the category of redefining the problem to match a potentially unsuitable evaluation measure---how can one get better ranks in the candidate-answer-ranking task.
For this reason, although there exist approaches that use the updated data, we do not report comparison to any of them.

\begin{figure*}[t]
  \centering\scriptsize
  \def\s{\hspace*{2ex}}
  \renewcommand*{\arraystretch}{1.1}
  \begin{tabular}{@{}c@{\s}c@{\,}p{0.4\linewidth}@{\,}c@{\,}p{0.4\linewidth}@{}}
    %
    % First row of examples
    %
    \multirow{6}{*}{\includegraphics[width=0.2\linewidth]{COCO_val2014_000000000136}}
    %% Q1
    & Q:
    & Are they adult giraffe?
    & Q:
    & Are there other animals? \\
    %
    % under image
    &
    & \rank[purple]{\tiny GT} \textcolor{purple}{Yes}
    &
    & \rank[purple]{\tiny GT} \textcolor{purple}{No} \\
    & \parbox[t]{2mm}{\multirow{4}{*}{\rotatebox[origin=c]{90}{\scriptsize Ranked Ans}}}
    & \rank{1} Yes the giraffe seems to be an adult
    & \parbox[t]{2mm}{\multirow{4}{*}{\rotatebox[origin=c]{90}{\scriptsize Ranked Ans}}}
    & \rank{1} No, there are no other animals \\
    & % over rot label
    & \rank{2} It seems to be an adult, yes
    & % over rot label
    & \rank{2} No other animals \\
    & % under rot label
    & \rank{3} The giraffe is probably an adult, it looks very big
    & % under rot label
    & \rank{3} There are no other animals around \\
    & % under rot label
    & \rank{4} Young adult
    & % under rot label
    & \rank{4} Don't see any animals \\[2ex]
    %
    % Second row of examples
    %
    \multirow{6}{*}{\includegraphics[width=0.2\linewidth]{COCO_val2014_000000000428}}
    %% Q1
    & Q:
    & Any candles on cake?
    & Q:
    & Is the cake cut? \\
    %
    % under image
    &
    & \rank[purple]{\tiny GT} \textcolor{purple}{Just a large ``number one''}
    & %\parbox[t]{2mm}{\multirow{4}{*}{\rotatebox[origin=c]{90}{\scriptsize Ranked Ans}}}
    & \rank[purple]{\tiny GT} \textcolor{purple}{No, but the boy has sure had his hands in it!} \\
    & \parbox[t]{2mm}{\multirow{4}{*}{\rotatebox[origin=c]{90}{\scriptsize Ranked Ans}}}
    & \rank{1} There are no candles on the cake
    & \parbox[t]{2mm}{\multirow{4}{*}{\rotatebox[origin=c]{90}{\scriptsize Ranked Ans}}}
    & \rank{1} No it's not cut \\
    & % over rot label
    & \rank{2} I actually do not see any candles on the cake
    & % over rot label
    & \rank{2} No the cake has not been cut \\
    & % under rot label
    & \rank{3} No , no candles
    & % under rot label
    & \rank{3} Nothing is cut \\
    & % under rot label
    & \rank{4} No candles
    & % under rot label
    & \rank{4} No, the cake is whole
  \end{tabular}
  \vspace*{-1.5ex}
  \caption{%
    \small%
    Example answers ``generated'' using the nearest-neighbours approach.
    For a given test question, a custom candidate set is constructed by choosing answers corresponding to the 100 closest (by correlation using A-Q) questions from the training data, and the best correlated answers to the given question returned.
  }
  \label{fig:nearest_neighbors}
  \vspace*{-5ex}
\end{figure*}

Although standard evaluation for \gls{VD} involves ranking the given candidate answers, there remains an issue of whether, given a question (relating to an image), the \gls{CCA} approach really ``answers'' it.
From one perspective, simply choosing from a given candidate set can seem a poor substitute for the ability to \emph{generate} answers, in the vein of \citet{Das_VisualDialog,massiceti2018}.
To address this, we construct a simple ``generative'' model using our learned projections between questions and answers (A-Q model, c.f. \cref{fig:nearest_neighbors}).
For a given question, we select the corresponding answers to the 100 nearest-neighbour questions using solely the train set and construct a custom candidate-answer set. We then compute their correlations with the given question, and sample the top-correlated answers as ``generated'' answers.
%\footnote{This only additionally requires disk-space to persist the training data---costing roughly \$0.25/GB}.
%
%Computing our heuristic for the validity of the top-ranked answers as before, with average variance (i) 0.018 and fraction correct (ii) 87.2\%, indicating reliable performance.

\vspace*{-1ex}
\section{Discussion}
\label{sec:discussion}
\vspace*{-1ex}
%% Dataset
We use the surprising equivalence from \cref{sec:experiments} as evidence of several issues with current approaches to \gls{VD}.
The biggest concern our evaluation, and a similar by \citep{anand2018blindfold}, reveals is that, for standard datasets in the community, \emph{visually grounded} questions can be answered ``well'', without referring to the visual stimuli.
This reveals an unwanted bias in the data, whereby correlations between question-answer pairs can be exploited to provide reasonable answers to visually-grounded questions.
Moreover, as indicated in \cref{fig:vd-fail}, the dataset also includes an implicit bias that any given question \emph{must} necessarily relate to a given image---as evidence by visually-unrelated questions getting visually-unrelated, but plausible answers.
A particularly concerning implication of this is that current approaches to \acrlong{VD}~\citep{Das_VisualDialog,visdial_rl,massiceti2018} may not actually be targetting the \emph{intended} task.

%% Metric
Our simple \gls{CCA} method also illustrates, that the standard evaluation used for \gls{VD} has certain shortcomings.
Principally, the use of ``candidate'' answers for each question, with a particular subset of them (1 in \textit{VisDial v0.9}, and K-human-derived weighted choices in v1.0) are deemed to be the ``ground-truth'' answers.
However, as we show in \cref{fig:qualitative-examples}, such an evaluation can still be insufficient to capture the range of all plausible answers.
The task of designing evaluations on the ``match'' of expected answers in for natural language, though, is fraught with difficulty, as one needs to account for a high degree of syntactic variability, with perhaps little semantic difference.

Responses to addressing the issues observed here, can take a variety of forms.
For the objective itself, one could alternately evaluate the effectiveness with which the dialogue enables a downstream task, as explored by some \citep{visdial_rl,de2017guesswhat,khani2018planning,LazaridouPB16a}.
Also, to address implicit biases in the dataset, one could adopt synthetic, or simulated, approaches, such as \citet{HermannHGWFSSCJ17}, to help control for undesirable factors.
Fundamentally, the important concern here is to evaluate visual dialogue on its actual utility---conveying information \emph{about the visual stimuli}---as opposed to surface-level measures of suitability.

%% Use simple baselines
And finally, we believe an important takeaway from our analyses is that it is highly effective to begin exploration with the simplest possible tools one has at one's disposal.
This is particularly apposite in the era of deep neural networks, where the prevailing attitude appears to be that it is preferable to start exploration with complicated methods that aren't well understood, as opposed to older, perhaps even \emph{less fashionable} methods that have the benefit of being rigorously understood.
Also, as shown in \cref{tab:compute-time}, choosing simpler methods can help minimise human effort and cost in terms of both compute and time, and crucially provide the means for cleaner insights into the problems being tackled.

\subsubsection*{Acknowledgements}
This work was supported by the ERC grant ERC-2012-AdG 321162-HELIOS, EPSRC grant Seebibyte EP/M013774/1, EPSRC/MURI grant EP/N019474/1, FAIR \emph{ParlAI} grant, and the Skye Foundation. We thank Abhishek Das for his comments and discussions on our manuscipt.

\bibliography{references}

\appendix
\vspace*{-1ex}
\glsresetall
\section{Multi-view Canonical Correlation Analysis}
\label{sec:app-cca}

Among several possible ways to formulate the \gls{CCA} objective for multiple variables, we choose the Frobenius norm-based objective of~\citet{hardoonMultiCCA2004}.
Let us assume that there are \(m\) views and \(\bfx_i \in \real^{n_i}\) represents an observation from the \(i\)\textsuperscript{th} view.
\(X_i \in \real^{n_i \times N}\) represents the column-wise stack of \(N\) observations from the \(i\)\textsuperscript{th} view.
The objective is to jointly learn projection matrices \(W_i \in \real^{n_i \times p}\) for all the \(m\) views such that the embeddings in the \(p (\leq n_i \forall i)\)-dimensional space are maximally correlated. This is achieved by optimising the following problem:
\begin{align*}
  &\min_{W_1, \dots, W_m}  \sum_{i,j=1, i \neq j}^m \norm{ X^\top_i W_i - X^\top_j W_j }^2_F
  \label{eq:app-mvcca} \numberthis \\
  &s.t.\;
    W_i^{\!\top}\! C_{ii} W_i = \mathbb{I},\quad\quad
    {{\bf w}_i^l}^{\!\top}\! C_{ij} {\bf w}_j^n \!=\! 0,\\
  & i,j=1,\ldots,m, i \!\neq\! j,
    l,n \!=\! 1, \ldots, p,\;l \!\neq\! n
\end{align*}
where \({\bf w}_i^l\) is the \(l\)\textsuperscript{th} column of \(W_i\), and \(C_{ij}\) is the correlation matrix between the \(i\)\textsuperscript{th} and \(j\)\textsuperscript{th} views, or \(C_{ij} = \frac{1}{N-1} X_iX_j^T\). \citet{bachkernelICA2002} show that optimising \cref{eq:app-mvcca} reduces to solving a generalised eigenvalue decomposition problem of the following form:
\begin{align*}
  A\bm{v} &= \lambda B\bm{v} \\
  \begin{pmatrix}
    C_{11}  & C_{12}  & \cdots & C_{1m} \\
    C_{21}  & C_{22}  & \cdots & C_{1m} \\
    \vdots  & \vdots & \ddots & \vdots \\
    C_{m1}  & C_{m2}  & \cdots & C_{mm}
  \end{pmatrix}
  \begin{pmatrix}
    \bm{v}_1 \\
    \bm{v}_2 \\
    \vdots \\
    \bm{v}_m
  \end{pmatrix}
  &= \lambda  \begin{pmatrix}
    C_{11}  & 0      & \cdots & 0\\
    0       & C_{22} & \cdots & 0\\
    \vdots  & \vdots & \ddots & \vdots \\
    0       & \cdots & \cdots & C_{mm}
  \end{pmatrix}
  \begin{pmatrix}
    \bm{v}_1 \\
    \bm{v}_2 \\
    \vdots \\
    \bm{v}_m
  \end{pmatrix}
\end{align*}
The top \(p\) (eigenvalue-sorted) eigenvectors \(\bm{v}_i \in \real^{n_i}\) are column-wise stacked to construct projection matrix \(W_i\) for view \(i \in \{1, \dots, m\}\).

\section{Rebuttal: ``Response to `Visual Dialogue without Vision or Dialogue' ''}
\label{sec:response}

In response to our original workshop submission at NeurIPS 2018, Das et. al.\ (2019) published a manuscript setting out two concerns, from their perspective, with our work.
Here, we provide a rebuttal for those concerns, noting that the authors seem to miss the primary implication drawn from our analysis, densely packed due to page limitations, that ranking candidate answers is an insufficient evaluation for this dataset and task combination.

To emphasize, with this work we are not advocating \gls{CCA} for \gls{VD}, but rather that when an extremely simple and common model does `quite well' compared to models with upwards of a few millions of parameters, requiring days for training, it points to potential issues with the setup, data and evaluation metrics.
We hope that these analyses encourage the community to take a step back and think carefully about the different aspects of the task.

%We believe that such analyses not just help us in designing better models, also help the research community in moving in the right direction saving tons of money and manpower, and in the end, solving the right problem.
%
% The leading question then is: How good is doing `quite well'?

\subsubsection*{Concern 1: Suitability of NDCG evaluation}
\label{sec:conc-ndcg}

The concern is that, contrary to what was implied in our original manuscript, the introduction of the NDCG metric has \emph{not} changed the nature of the visual dialogue task to an explicit classification task.

We characterised the task as having become an explicit classification task primarily because, at time of publication of our original manuscript, the only models using \textit{v1.0} that were publicly available on the \href{https://visualdialog.org/challenge/2018}{Visual Dialog Challenge 2018} leaderboard, were the ``discriminative'' models.
Moreover, the leaderboard did not make a distinction between ``generative'' and ``discriminative'' models, which is of concern as it can encourage the community to embrace the ``discriminative'' task in an approach to improve position on the leaderboard.

Since then, the challenge task has been updated to include an FAQ about performance differences between the two types of models, and there have also been publicly released ``generative'' models employing the \textit{v1.0} dataset (with NDCG evaluation), which we compare to the \gls{CCA} model in \cref{tab:ndcg-comparison}.
Although the NDCG performance of \gls{CCA} is below that of the \gls{SOTA}, as explained in the next section, lower-than-\gls{SOTA} performance on an evaluation task that involves matching answers in a candidate set does not necessarily imply that the model performs poorly on the task.

\begin{table}[t]
  \centering \footnotesize
  \newcommand{\citelink}[2]{#2, \citep{#1}}
  \def\hre{\citelink{Das_VisualDialog}{HRE-QIH-G}}
  \def\lfQIH{\citelink{Das_VisualDialog}{LF-QIH-G}}
  \def\aq{\gls{CCA} A-Q}
  \def\s{\hspace*{5pt}}
  % \vspace*{-2\baselineskip}
  \caption{%
    Results for \gls{SOTA} vs.\ \gls{CCA} on \textit{VisDial v1.0} including the NDCG metric.
  }
  \label{tab:ndcg-comparison}
  \scalebox{1.0}{%
  \begin{tabular}{@{}>{\footnotesize}c@{\s}l@{\s}c@{\s}c@{\s}c@{\s}c@{\s}c@{\s}c@{}}
    \toprule
    Model               & I/QA features      & MR    & R@1   & R@5   & R@10  & MRR    & NDCG   \\
    \midrule
    \hre                & VGG-16/learned     & 18.78 & 34.78 & 56.18 & 63.72 & 0.4561 & 0.5245 \\
    \lfQIH              & VGG-16/learned     & 18.81 & 35.08 & 55.92 & 64.02 & 0.4568 & 0.5121 \\
    \cmidrule{2-8}
    \multirow{2}*{\aq}  & GloVe              & 16.60 & 16.10 & 39.38 & 54.68 & 0.2824 & 0.3504 \\
                        & FastText           & 17.07 & 16.18 & 40.18 & 55.35 & 0.2845 & 0.3493 \\
    \bottomrule
  \end{tabular}}
  \vspace*{-1.5\baselineskip}
\end{table}

\vspace*{-1.5ex}
\subsubsection*{Concern 2: Comparison to proposed \gls{CCA} baseline}
\label{sec:conc-metric}
\vspace*{-1ex}
The concern is that we do not approach state-of-the-art on \emph{all} metrics, but only on mean rank (MR).

We did make this clear and explicit in the paper (c.f. discussion about \cref{tab:cca-analysis} in \cref{sec:experiments}).
To further reinforce this, we have added explicit references to MR in the abstract and closing of the introduction.

Regarding the discrepancies in recall, MRR, and NDCG, however, the implication we wish to draw with regard to doing `quite well' is that a metric (or set of metrics) that targets the membership of a predicted answer in a set (a singleton set in the case of MR/R@/MRR, and a slightly larger set for NDCG) does not appear to be well correlated with the ability to faithfully answer a given question, which is something we emphasise qualitatively in \cref{fig:qualitative-examples}.
Note that these results were \emph{anti}-cherry-picked---we deliberately chose examples with \emph{poor} ground-truth rank, and observed what the top-ranked answers were.

In analysing the reason behind this surprising performance, we employ a simple heuristic to characterise the correlation of the question to the different answers in the candidate set, computing an automatic binary partition on them.
We use the Otsu threshold, but others can be used and show similar results.
Subsequently, we look at the tightness of the better-ranked cluster, and how often this cluster includes the ``ground-truth'' answer.
The results indicate that the correlations (in range [-1, 1]) are quite closely packed, and almost all of the time, the intended ``ground-truth'' answer is in that cluster and close in correlation to the top answers predicted by the \gls{CCA} method---strongly implying that the answers in the better-ranked cluster (including the ``ground-truth'') are effectively an \emph{equivalence class} in terms of being able to answer a given question.
This would imply that the purported rank of the ``ground-truth'' answer could be arbitrary within that class, hence highlighting the unsuitability of the ranking metric in this instance.
This effectively implies that \emph{doing poorly on the given metrics does not imply poor performance on the actual underlying task.}

The authors also cite a lack of comparison to nearest-neighbour baselines as in~\cite{Das_VisualDialog}, to ablative variants of \gls{SOTA} models (e.g. when the image and/or history is removed), and to the \gls{CCA} approach run with the same pre-trained feature extractors as \gls{SOTA} models.
To comment on the nearest-neighbour baselines, they are indeed useful comparisons and we have included the results of own implementation of them in \cref{tab:baseline-comparison}. We observe that \gls{CCA} is still superior in MR, and additionally in computation and storage requirements compared to the nearest-neighbour approach which requires the train questions and answers (and images) at test time.
Against ablative model variants, \gls{CCA}'s MR is comparable, and in most cases actually out-performs ablations of the \gls{SOTA} models.

Regarding our use of ResNet and FastText rather than VGG and learned word embeddings, as we mentioned above, the paper's primary focus is to comment on the visual dialogue task/evaluations as they are currently framed rather than to explore architectural effects. Nevertheless, for a complete comparison we have run the \gls{CCA} method using VGG and GloVe, which we also include in \cref{tab:ndcg-comparison}, and compare with in \cref{tab:baseline-comparison}.
Using GloVe embeddings in fact improves our results.

\begin{table}[h!]
  \centering \footnotesize
  \def\s{\hspace*{5pt}}
  \vspace*{-1.5\baselineskip}
  \caption{%
    Results for baseline from \citet{Das_VisualDialog} vs.\ \gls{CCA} on \textit{v0.9}.
  }
  \label{tab:baseline-comparison}
  \scalebox{1.0}{%
  \begin{tabular}{@{}>{\footnotesize}c@{\s}l@{\s}c@{\s}c@{\s}c@{\s}c@{\s}c@{}}
    \toprule
    Model               & I/QA features      & MR    & R@1   & R@5   & R@10  & MRR    \\
    \midrule
    \gls{CCA} A-Q       & GloVe              & 15.86 & 16.93 & 44.83 & 58.44 & 0.3044 \\
    \gls{CCA} A-QI (Q)  & VGG-16/GloVe       & 26.03 & 12.24 & 30.96 & 42.63 & 0.2237 \\
    \cmidrule{1-7}
    NN-A-Q            & GloVe              & 19.67 & 29.88 & 47.07 & 55.44 & 0.3898 \\
    NN-A-QI           & VGG-16/GloVe       & 20.14 & 29.93 & 46.42 & 54.76 & 0.3873 \\
    \bottomrule
  \end{tabular}}
  \vspace*{-1\baselineskip}
\end{table}

\subsubsection*{References}
Abhishek Das, Devi Parikh, Dhruv Batra. Response to ``Visual Dialogue without Vision or Dialogue''\\ \rule{1em}{0em}(Massiceti et al., 2018). arXiv preprint arXiv:1901.05531, 2019.

\end{document}

%% file: maths-basic.tex
\usepackage{scalerel}
\usepackage{mleftright}
\usepackage{dsfont}
\usepackage{bm}

\newcommand{\real}{\mathbb{R}}

%% file: main.bbl
\begin{thebibliography}{21}
\providecommand{\natexlab}[1]{#1}
\providecommand{\url}[1]{\texttt{#1}}
\expandafter\ifx\csname urlstyle\endcsname\relax
  \providecommand{\doi}[1]{doi: #1}\else
  \providecommand{\doi}{doi: \begingroup \urlstyle{rm}\Url}\fi

\bibitem[Anand et~al.(2018)Anand, Belilovsky, Kastner, Larochelle, and
  Courville]{anand2018blindfold}
Ankesh Anand, Eugene Belilovsky, Kyle Kastner, Hugo Larochelle, and Aaron
  Courville.
\newblock {Blindfold Baselines for Embodied QA}.
\newblock \emph{arXiv preprint arXiv:1811.05013}, 2018.

\bibitem[Arora et~al.(2017)Arora, Liang, and Ma]{aurora2017simple}
Sanjeev Arora, Yingyu Liang, and Tengyu Ma.
\newblock A simple but tough-to-beat baseline for sentence embeddings.
\newblock 2017.

\bibitem[Bach and Jordan(2002)]{bachkernelICA2002}
Francis.~R Bach and Michael.~I. Jordan.
\newblock Kernel independent component analysis.
\newblock \emph{Journal of Machine Learning Research}, 2002.

\bibitem[Bojanowski et~al.(2017)Bojanowski, Grave, Joulin, and
  Mikolov]{bojanowski2017enriching}
Piotr Bojanowski, Edouard Grave, Armand Joulin, and Tomas Mikolov.
\newblock Enriching word vectors with subword information.
\newblock \emph{Transactions of the Association for Computational Linguistics},
  2017.

\bibitem[Das et~al.(2017{\natexlab{a}})Das, Kottur, Gupta, Singh, Yadav, Moura,
  Parikh, and Batra]{Das_VisualDialog}
Abhishek Das, Satwik Kottur, Khushi Gupta, Avi Singh, Deshraj Yadav,
  Jos\'e~M.F. Moura, Devi Parikh, and Dhruv Batra.
\newblock {V}isual {D}ialog.
\newblock In \emph{{IEEE} Conference on Computer Vision and Pattern
  Recognition}, June 2017{\natexlab{a}}.

\bibitem[Das et~al.(2017{\natexlab{b}})Das, Kottur, Moura, Lee, and
  Batra]{visdial_rl}
Abhishek Das, Satwik Kottur, Jos\'e~M.F. Moura, Stefan Lee, and Dhruv Batra.
\newblock Learning cooperative visual dialog agents with deep reinforcement
  learning.
\newblock In \emph{Proc. Int. Conf. Computer Vision}, 2017{\natexlab{b}}.

\bibitem[De~Vries et~al.(2017)De~Vries, Strub, Chandar, Pietquin, Larochelle,
  and Courville]{de2017guesswhat}
Harm De~Vries, Florian Strub, Sarath Chandar, Olivier Pietquin, Hugo
  Larochelle, and Aaron~C Courville.
\newblock {GuessWhat?! Visual object discovery through multi-modal dialogue}.
\newblock In \emph{{IEEE} Conference on Computer Vision and Pattern
  Recognition}, 2017.

\bibitem[Gong et~al.(2014)Gong, Ke, Isard, and Lazebnik]{gongmultiCCA14}
Yunchao Gong, Qifa Ke, Michael Isard, and Svetlana Lazebnik.
\newblock A multi-view embedding space for modeling internet images, tags, and
  their semantics.
\newblock \emph{Int. Journal of Computer Vision}, 2014.

\bibitem[Hardoon et~al.(2004)Hardoon, Szedmak, and
  Shawe-taylor]{hardoonMultiCCA2004}
David~R. Hardoon, Sandor~R. Szedmak, and John~R. Shawe-taylor.
\newblock Canonical correlation analysis: An overview with application to
  learning methods.
\newblock \emph{Neural Computation}, 2004.

\bibitem[He et~al.(2016)He, Zhang, Ren, and Sun]{He_2016deep}
Kaiming He, Xiangyu Zhang, Shaoqing Ren, and Jian Sun.
\newblock Deep residual learning for image recognition.
\newblock In \emph{{IEEE} Conference on Computer Vision and Pattern
  Recognition}, 2016.

\bibitem[Hermann et~al.(2017)Hermann, Hill, Green, Wang, Faulkner, Soyer,
  Szepesvari, Czarnecki, Jaderberg, Teplyashin, Wainwright, Apps, Hassabis, and
  Blunsom]{HermannHGWFSSCJ17}
Karl~Moritz Hermann, Felix Hill, Simon Green, Fumin Wang, Ryan Faulkner, Hubert
  Soyer, David Szepesvari, Wojciech~Marian Czarnecki, Max Jaderberg, Denis
  Teplyashin, Marcus Wainwright, Chris Apps, Demis Hassabis, and Phil Blunsom.
\newblock {Grounded Language Learning in a Simulated 3D World}.
\newblock \emph{CoRR}, abs/1706.06551, 2017.
\newblock URL \url{http://arxiv.org/abs/1706.06551}.

\bibitem[Hotelling(1936)]{hotelling1936}
Harold Hotelling.
\newblock Relations between two sets of variates.
\newblock \emph{Biometrika}, 1936.

\bibitem[Kettenring(1971)]{kettenringMultiCCA1971}
J.~R. Kettenring.
\newblock Canonical analysis of several sets of variables.
\newblock \emph{Biometrika}, 1971.

\bibitem[Khani et~al.(2018)Khani, Goodman, and Liang]{khani2018planning}
Fereshte Khani, Noah~D. Goodman, and Percy Liang.
\newblock Planning, inference and pragmatics in sequential language games.
\newblock \emph{CoRR}, abs/1805.11774, 2018.
\newblock URL \url{http://arxiv.org/abs/1805.11774}.

\bibitem[Lazaridou et~al.(2016)Lazaridou, Pham, and Baroni]{LazaridouPB16a}
Angeliki Lazaridou, Nghia~The Pham, and Marco Baroni.
\newblock Towards multi-agent communication-based language learning.
\newblock \emph{CoRR}, abs/1605.07133, 2016.
\newblock URL \url{http://arxiv.org/abs/1605.07133}.

\bibitem[Lu et~al.(2017)Lu, Kannan, Yang, Parikh, and Batra]{lu2017best}
Jiasen Lu, Anitha Kannan, Jianwei Yang, Devi Parikh, and Dhruv Batra.
\newblock Best of both worlds: Transferring knowledge from discriminative
  learning to a generative visual dialog model.
\newblock In \emph{Advances in Neural Information Processing Systems}, pages
  314--324, 2017.

\bibitem[Massiceti et~al.(2018)Massiceti, Siddharth, Dokania, and
  Torr]{massiceti2018}
Daniela Massiceti, N.~Siddharth, Puneet~K. Dokania, and Philip~H.S. Torr.
\newblock {FlipDial: A Generative Model for Two-Way Visual Dialogue}.
\newblock In \emph{{IEEE} Conference on Computer Vision and Pattern
  Recognition}, 2018.

\bibitem[Otsu(1979)]{otsu1979threshold}
Nobuyuki Otsu.
\newblock A threshold selection method from gray-level histograms.
\newblock \emph{IEEE transactions on systems, man, and cybernetics}, 9\penalty0
  (1):\penalty0 62--66, 1979.

\bibitem[Weizenbaum(1966)]{weizenbaum1966eliza}
Joseph Weizenbaum.
\newblock {ELIZA}—{A} computer program for the study of natural language
  communication between man and machine.
\newblock \emph{Communications of the ACM}, 1966.

\bibitem[Winograd(1971)]{winograd1971procedures}
Terry Winograd.
\newblock Procedures as a representation for data in a computer program for
  understanding natural language.
\newblock Technical report, Massachusetts Institiue of Technology, Cambridge,
  MA, 1971.

\bibitem[Wu et~al.(2017)Wu, Wang, Shen, Reid, and van~den Hengel]{wu2017you}
Qi~Wu, Peng Wang, Chunhua Shen, Ian Reid, and Anton van~den Hengel.
\newblock Are you talking to me? reasoned visual dialog generation through
  adversarial learning.
\newblock \emph{arXiv preprint arXiv:1711.07613}, 2017.

\end{thebibliography}
